\documentclass[runningheads]{llncs}

\usepackage[mobile]{eccv}

\usepackage{eccvabbrv}

\usepackage{graphicx}
\usepackage{booktabs}

\usepackage[accsupp]{axessibility}  

\usepackage[pagebackref,breaklinks,colorlinks,citecolor=eccvblue]{hyperref}

\usepackage{orcidlink}
\usepackage{booktabs}       
\usepackage{amsfonts}       
\usepackage{nicefrac}       
\usepackage{microtype}      
\usepackage{multirow}
\usepackage{wrapfig}
\usepackage{float}
\usepackage{array}
\usepackage{tabu}
\usepackage{amssymb}
\usepackage{tikz}
\usepackage[accsupp]{axessibility}
\usepackage{longtable}
\usepackage{makecell}
\usepackage[export]{adjustbox}
\usepackage{pifont}
\newcommand{\cmark}{\ding{51}}%
\newcommand{\xmark}{\ding{55}}%
\usetikzlibrary{decorations}
\usetikzlibrary{shapes}
\usetikzlibrary{3d}
\usetikzlibrary {shapes.geometric} 
\usetikzlibrary{shapes.symbols}
\usetikzlibrary{shapes.multipart}
\usepackage[acronym]{glossaries}
\usepackage{siunitx} 
\makeglossaries
\newacronym{vig}{ViG}{Vision GNN}
\newacronym{pvig}{PViG}{Pyramid ViG}

\newacronym{vit}{ViT}{Vision Transformer}
\newacronym{gnn}{GNN}{Graph Neural Network}
\newacronym{gcn}{GCN}{Graph Convolutional Networks}
\newacronym{gsat}{GSAT}{Graph Stochastic Attention}
\newacronym{wignet}{WiGNet}{Windowed Vision Graph Neural Network}
\newacronym{xai}{xAI}{Explainable AI}
\newacronym{iwivig}{i-WiViG}{Interpretable Window Vision GNN}
\newacronym{ig}{IG}{Integrated Gradients}
\newacronym{cnn}{CNN}{Convolutional Neural Network}

\newcommand{\methodname}{i-WiViG}

\begin{document}

\title{i-WiViG: Interpretable Window Vision GNN} 

%
\titlerunning{i-WiViG: Interpretable Window Vision GNN}


\author{Ivica Obadic\inst{1,2}\orcidlink{0000-0003-4403-2170} \and
Dmitry Kangin\inst{3, 4}\orcidlink{0000-0001-9769-7585} \and
Adrian Höhl\inst{1}\orcidlink{0000-0003-3380-4489}
\and
Dario Oliveira\inst{5}\orcidlink{0000-0002-0674-5332}
\and \quad
Plamen P Angelov\inst{3}\orcidlink{0000-0002-5770-934X}
\and
Xiaoxiang Zhu\inst{1,2}\orcidlink{0000-0001-5530-3613}}

\authorrunning{I. Obadic et al.}

\institute{Technical University of Munich \and
Munich Center for Machine Learning
\and
University of Lancaster \and
University of Manchester \and 
Getulio Vargas Foundation
}

\maketitle

\begin{abstract}
Vision graph neural networks have emerged as a popular approach for modeling the global and spatial context for image recognition. However, a significant drawback of these methods is that they do not offer an inherent interpretation of the relevant spatial interactions for their prediction. We address this problem by introducing \textbf{\methodname}, an approach that enables interpretable model reasoning based on a sparse subgraph in the image. \methodname\ is based on two key postulates: 1) constraining the graph nodes' receptive field to disjoint local windows in the image, and 2) an inherently interpretable graph bottleneck with learnable sparse attention that identifies the relevant interactions among the local image windows. We evaluate our approach on both scene classification and regression tasks using natural and remote sensing imagery. Our results, supported by quantitative and qualitative evidence, demonstrate that the method delivers semantic, intuitive, and faithful explanations through the identified subgraphs. Furthermore, extensive experiments confirm that it achieves competitive performance to its black-box counterparts, even on datasets exhibiting strong texture bias. The implementation is available on \url{https://github.com/zhu-xlab/i-WiViG}.
\end{abstract}
\section{Introduction}
\begin{figure}[h!]
\centering{
\resizebox{0.95\textwidth}{!}{
\begin{tikzpicture}[domain=0:15]
\node at (0, 1.85) {\small{\makecell{Overlapping\\receptive fields}}};
\node at (2.3, 1.85) {\small{\makecell{Spatially incon-\\sistent graph}}};

\draw[blue, thick, rounded corners, dashed] (-1.1, -0.5) rectangle (5.6,2.75);

\node[blue,font=\bf] at (2, 2.5) {Vision GNNs};
\node at (0,0.45)  (twonodes1) {\includegraphics[width=2cm]{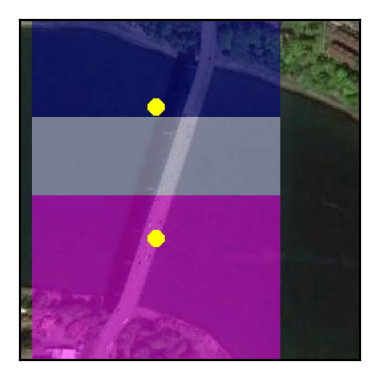}};
\node at (2.25,0.45)  (receptionfield1) {\includegraphics[width=2cm]{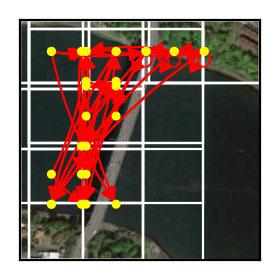}};

\node at (4.4,0.45)  (explanation1b) {\makecell{No  inherent \\ explanations}};

 \begin{scope}[shift={(7,3.25)}]
 
\draw[blue, thick, rounded corners, dashed] (-1.1, -3.75) rectangle (5.6,-0.5);

\node at (0, -1.4) {\small{\makecell{Disjoint\\ nodes}}};
\node at (2.25, -1.4) {\small{\makecell{Sparse\\ subgraphs}}};
\node at (4.4, -1.4) {\small{\makecell{Faithful \\ explanations}}};

Disjoint nodes, Bottom 2: Intuitive Explanation, Bottom 3: Faithful explanation

\node at (0,-2.75)  (twonodes2) {\includegraphics[width=2cm]{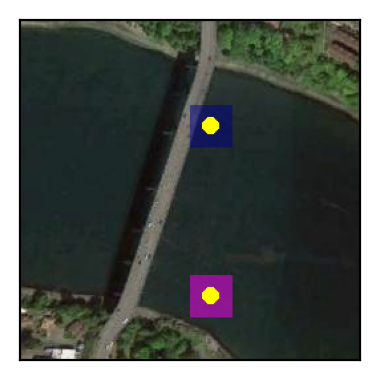}};

\node at (2.25,-2.75) (receptionfield2) {\includegraphics[width=2cm]{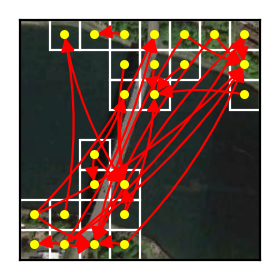}};

\node at (4.5,-2.75) (globalconnections) {\includegraphics[width=2.1cm]{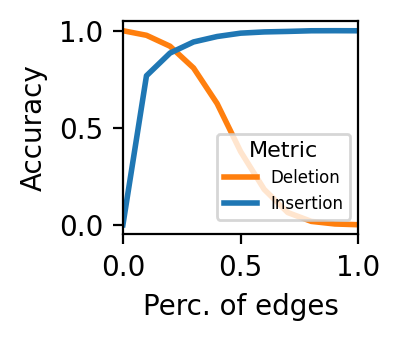}};

\node[blue, font=\bf] at (2.1, -0.75) {\methodname};
\end{scope}

\node at (4, -0.78) {\small{{\color[RGB]{211,211,255}{$\blacksquare$}} Node overlap}};
\node at (0, -0.75) {\small{{\color[RGB]{0, 0, 139}{$\blacksquare$}} Node 1}};
\node at (1.8, -0.75) {\small{{\color[RGB]{255, 0, 255}{$\blacksquare$}} Node 2}};

\end{tikzpicture}
}
}
\caption{\textit{Left}: Vision GNNs produce graph nodes with large and overlapping receptive fields. 
This prevents the understanding of the relevant long-range interactions among fine-grained objects in the image.
\textit{Right}: Our \methodname\ method constrains the receptive fields of the graph nodes to a local window in the image. Further, it introduces an inherently interpretable graph bottleneck based on sparse attention that identifies the relevant long-range interactions among the image windows. Unlike Vision GNNs, \methodname\ explanatory subgraphs reveal semantic relations in the image that faithfully explain the model decisions.}
\label{teaser_figure}
\end{figure}

Traditional computer vision methods, such as \glspl{cnn}, often exhibit a bias toward capturing local image texture rather than object composition and structure \cite{geirhos2018imagenet, sabour2017dynamic}. To overcome this limitation, recent approaches
such as \glspl{vit} \cite{dosovitskiy2021an} or \acrfull{vig} \cite{han2022vision} have shifted the focus from modeling local contexts to encoding the global interactions within an image. While \glspl{vit} represent an image as a sequence, the \gls{vig} models \cite{han2022vision, han2023vision, munir2023mobilevig, munir2024greedyvig, spadaro2024wignet} encode it into a graph, offering a more flexible structure capable of modeling complex image objects with irregular shapes. Furthermore, by connecting only a subset of nodes, the \gls{vig} models encode spatial dependencies with fewer parameters than standard ViTs, which construct a dense graph between the image patches and utilize self-attention to model the pairwise spatial interactions.

Capturing the topological relationships is critical for tasks such as remote sensing image recognition, where images can cover large geographic areas and include objects of varying scales and geometries \cite{10777297, HU2024465, liu2021self}. The capabilities of the \glspl{vig} in these contexts 
make them an attractive choice for processing remote sensing imagery, and they are increasingly adapted for applications such as land-cover change detection and classification \cite{song2024context, you2023crossed, zhang2024hcgnet, shou2023graph, LI2024114290} or anomaly detection after a natural hazard \cite{chen2024hierarchical}. However, a significant limitation of \glspl{vig} is their black-box nature, as they lack a mechanism to unveil the relevant connections in the image that drive the model predictions. This arises from the following two key issues:
\begin{itemize}
    \item \textbf{Overlapping nodes}: The different \gls{vig} models produce large and overlapping receptive fields. This is visualized in the left part of Figure \ref{teaser_figure} for the \gls{wignet} model \cite{spadaro2024wignet}, where the receptive fields of two nodes that are located around the top and middle of the bridge display a large overlap. Both nodes include land, at least half of the bridge, and water. Modeling the global dependencies on such a graph disables the intuitive understanding of the relevant objects and their long-range interactions in the image.
    \item \textbf{Explanation methods for image graphs:} While there exist a large number of methods providing an inherent interpretation into the workings of the \glspl{gnn} \cite{kakkad2023surveyexplainabilitygraphneural}, they typically operate on data having natural graph structure. Yet, their performance is not evaluated on image graphs, which usually contain noisy and redundant edges \cite{han2023vision}. Thus, a straightforward application of these approaches to image graphs can fail to accurately identify the relevant spatial interactions in the image.
\end{itemize}

This work addresses the above shortcomings by proposing \methodname, an inherently interpretable \gls{vig} approach that outputs an explanation highlighting the relevant subgraph in the input image for the model prediction. We achieve this by injecting the following constraints in the model architecture: 1) limiting the receptive field of a graph node to a small and contiguous window in the image, and 2) introducing an inherently interpretable graph bottleneck based on sparse edge attention to encode the global relations in the image. The first step produces graph nodes with disjoint receptive fields. Next, the sparse edge attention in the graph bottleneck automatically filters the relevant connections. As a result, our approach generates a sparse subgraph during inference time, automatically identifying the relevant spatial interactions within the image. This is visualized in the right part of Figure \ref{teaser_figure}, which illustrates that the identified subgraph faithfully emphasizes the importance of the long-range connections between both sides of the land for predicting the bridge class.

In summary, this work introduces \methodname, an inherently interpretable \gls{vig} approach, by providing the following main contributions: 
\begin{itemize}
    \item An interpretable reasoning mechanism that models spatial interactions within the image, enabled by a visual encoder that maps local image windows into disjoint graph nodes and a sparse graph attention bottleneck that identifies the relevant subgraph for the prediction.
    \item We qualitatively and quantitatively compare our identified subgraphs with those of the existing \gls{gnn} explanation methods, and demonstrate that it produces intuitive explanations with improved faithfulness over the benchmarks.
    \item Finally, we show that, despite imposing the constraint through disjoint nodes and the graph attention bottleneck, our model delivers performance on par with black-box counterparts, while preserving interpretability even on datasets with pronounced texture bias.
\end{itemize}
\section{Related Work}
\paragraph{Graph Neural Networks for Computer Vision}
The \gls{vig} approaches \cite{han2022vision, munir2023mobilevig, munir2024greedyvig, spadaro2024wignet, 10845790, parikh2025clustervig, han2023vision, gedik2025attentionvig} combine \gls{cnn} layers for extracting local textural features and graph message passing for encoding long-range relations between these features. They first extract graph nodes encoding rectangular image patches with a series of \gls{cnn} layers and add edges among them based on a similarity metric.
The pioneering work of Han et al. \cite{han2022vision} connects a node with its $k$-nearest neighbours based on their embedding similarity in the feature map. This can be computationally expensive as it requires computing the similarity between all pairs of graph nodes.
The follow-up works generally focus on three main areas: reducing the computational bottleneck of the $k$-nearest neighbors search \cite{munir2023mobilevig, munir2024greedyvig, spadaro2024wignet, 10845790, parikh2025clustervig}, improving the image-to-graph graph representation \cite{han2023vision}, and the aggregation of the node neighbourhood \cite{gedik2025attentionvig}. To tackle the $k$-NN computational bottleneck, researchers have proposed constructing the graph based on the node positions in the feature map \cite{munir2023mobilevig, munir2024greedyvig}, constraining the neighbour search to a grouped image segment, such as a rectangular local window \cite{spadaro2024wignet} or cluster \cite{parikh2025clustervig}, and employing a similarity threshold to select the optimal number of neighbours \cite{10845790}. The graph representation problem is addressed in \cite{han2023vision} by encoding the image into a hypergraph to capture high-order interactions within an image. Further, \cite{gedik2025attentionvig} proposes an attention-based node-neighbor feature aggregation method to dynamically weight the contribution of the node's neighbours.
Despite their impressive performance on computer vision tasks, a key problem with the aforementioned \gls{vig} approaches is their lack of interpretability, as they are unable to unveil the relevant spatial interactions for the model predictions.
\paragraph{GNN Explainability} The existing methods for \gls{gnn} explainability aim to identify the relevant edges and nodes in the graph that contribute to the outcome. These methods are commonly categorized into \textit{post hoc}  and \textit{ante hoc} methods \cite{kakkad2023surveyexplainabilitygraphneural}. 
The \textit{post hoc} methods aim at explaining a trained graph neural network. One of the most popular graph explanation methods in this category is the GNNExplainer \cite{ying2019gnnexplainer}, which identifies the relevant subgraph by maximizing mutual information between the predicted label distribution and the explanatory subgraph. \cite{yuan2020xgnn} instead takes a holistic approach towards \textit{post hoc} interpretability, generating the largest output inputs of a given size for each class. Another work \cite{lin2022orphicx} creates a causal explanation by maximizing the information flow measurements. However, a main drawback of the \textit{post hoc} approaches is that they usually do not come with strict guarantees for the fidelity of the explanations to the actual working of the model on which they operate \cite{yuan2022explainability,scafarto2024augment}. 
The nascent field of \textit{ante hoc} graph interpretation methods aims to propose architectures that, by design, propagate attribution to nodes and edges as part of the decision-making process. For instance, the \gls{gsat} approach \cite{miao2022interpretable} by Miao et al. constrains the model prediction to rely on the automatically learned edge weights and uses the notion of stochasticity to ensure that the edges relevant to the inference have a high weight. An example of another method is given in \cite{scafarto2024augment}, which studies the potential of data augmentation to identify the parts of the graph that contribute to the decision. Further, \cite{chen2024interpretable} uses a combinatorial optimization technique, called multilinear extension, to find out the subgraphs that matter for the decision-making process. 

\section{Preliminaries}

We define a graph $G(V, E)\in \mathbb{G}$, where $V$ are the vertices and $E$ are the edges of the graph, and $\mathbb{G}$ is the set of possible graphs. We consider the task of prediction of label mapping from the graph to the vector labels $y (\cdot): \mathbb{G} \rightarrow \mathbb{L}$, where $\mathbb{L}\subset \mathbb{R}^{k_{\mathbb{L}}}$ and $k_{\mathbb{L}}$ is the size of the output vector. 

\subsection{\acrfull{vig}}
\label{sec:preelminiaries_vig_models}
\paragraph{Vision GNN} Our work builds upon the \acrfull{vig} \cite{han2022vision} architecture, which leverages graph representation learning from patchwise image embeddings. They consist of the following three steps: (1) extracting graph nodes from the input image with the stem operation, (2) constructing the graph adjacency matrix, and (3) performing graph message passing to derive the outcome. The stem operation applies a series of convolutional layers that transform the input image into a high-level feature map. The pixels in this feature map encode rectangular patches in the input image and are used as graph nodes. Next, the adjacency matrix is created by connecting a node to its $k$ nearest neighbours based on their embedding similarity in the feature map.  \gls{vig} offers two architectures, namely \textit{isotropic} and \textit{pyramid}. The isotropic version maintains the number of nodes and their feature size. In contrast, the pyramid architecture alternates between graph layers that exchange global information and \gls{cnn} layers that downsample the spatial size of the feature map while increasing the dimensionality of the node embeddings. All \gls{vig} models use a \gls{cnn} layer with a kernel size of 3 and a stride of 2 for the downsampling, which introduces additional overlap between neighbouring graph nodes. Hence, they yield a global receptive field of the graph nodes dispersed across the different parts of the image. 

The \gls{vig} network defines the \textit{Grapher} module that uses the graph convolutional layers of the following form: 
\begin{equation}
    G'  = \operatorname{Update} (\operatorname {Aggregate} (G, \mathbf{w}_{a}), \mathbf{w}_u),
\end{equation}
where $\operatorname {Update}$ and $\operatorname{Aggregate}$ are the update and aggregation operators parameterised with the weights $\mathbf{w}_{a}$ and $\mathbf{w}_{u}$ respectively,  $G$ is an input graph and $G'$ is the output graph. To increase the feature diversity, the Grapher module introduces a linear layer before and after the graph convolution and, similar to the \gls{vit} model \cite{dosovitskiy2021an}, processes the features of each node in a feed-forward network. To enable training of deep architectures, the $k$-neighbours adjacency matrix is dynamically computed based on the output of the Grapher layer. 

\paragraph{\acrfull{wignet}}
\label{sec:wignet_description}
In contrast to the other \gls{vig} approaches, the \gls{wignet} model \cite{spadaro2024wignet} introduced by Spadaro et al. limits the graph node receptive fields to a smaller region in the image. This is achieved by grouping the patches from the stem operation into windows and afterward computing the $k$-nearest neighbor graph and applying the above Grapher layer from \gls{vig}, locally, within each window. The \gls{wignet} follows the pyramid architecture, that downsamples the feature map with the same convolution operation as the other \glspl{vig} after every stage of graph processing. This hierarchically increases the local windows of the graph nodes as the image traverses through the network.

\subsection{Inherently-Interpretable \glspl{gnn}}
\label{sec:preelminiaries_gsat}
The inherently interpretable \gls{gnn} models are often based on the information bottleneck objective, which aims to identify an explanation subgraph such that the mutual information between the subgraph and the original input graph is maximized. One prominent method in this category is the \acrfull{gsat} approach \cite{miao2022interpretable}, which is a self-interpretable graph neural network that learns to automatically select the relevant subgraph during the model inference. Along with the model prediction, \gls{gsat} outputs an attention over edge weights which form the explanation subgraph.
To identify the subgraph, the \gls{gsat} approach learns an attention distribution $\alpha$ over the graph edges by injecting stochasticity during training. In detail, the \gls{gsat} approach first estimates the probability $\alpha_{uv}$ of sampling the edge $(u, v)$ that is afterward used as the weight for the edge in the forward pass. The edge embedding is constructed by concatenating the embeddings of the node $u$ and the node $v$, and it is passed through a series of FFN layers that output $\alpha_{uv}$. Next, during training, a Bernoulli sampling is performed for each edge $(u, v)$ using $\alpha_{uv}$ as a sampling probability. This forces the model to assign high weights to the relevant edges, so that these edges are more likely to be selected in the resulting subgraph. Thus, the highest $\alpha_{uv}$ values indicate the critical edges for the model prediction. In practice, the \gls{gsat} approach is implemented by injecting a random noise during training to the estimated edge weights and by introducing hyperparameter $r$ that regularizes the edge weights with the following loss function:  $\mathcal{L} =  \mathcal{L}_{\text{task}} + \mathcal{L}_{\alpha}$, with 

\begin{equation}
\label{eq:gsat}
\mathcal{L}_{\alpha} = \sum_{i=1}^{|E|} \alpha_i \log \frac{\alpha_i}{r} + (1 - \alpha_i) \log \frac{1 - \alpha_i}{1 - r}
\end{equation}

We denote the edge attention weights as $\alpha_i$ for every $e_i \in E, i \in [1, |E|]$. The hyperparameter $r$ controls the distribution of the edge attention and makes $\mathcal{L}_{\alpha}$ term minimal when all edge weights are equal to $r$. At the same time, in case a subset of the edges bears the critical information for the model prediction, the task loss term will force those critical edges to have weights higher than $r$. Hence, the edges with the highest weights can be used to uncover the critical subgraph.  


\section{\methodname}
\label{sec:method}
\begin{figure}
\centering{
\resizebox{1.01\textwidth}{!}{
\begin{tikzpicture}[inner sep=0pt]

\node at (-1.25,-0.55) (input) {\includegraphics[width=1.95cm]{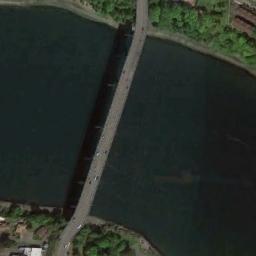}};

\node[draw, trapezium, rotate=-90, fill=gray!15, minimum width=1cm, minimum height=0.5cm, trapezium stretches=true] at (0.5,-0.55) (bagnet) {\makecell{\small 1.Window encoder}};

\node at (2.3, -0.55) (feature_map) {\includegraphics[width=2.2cm, height=2.2cm]{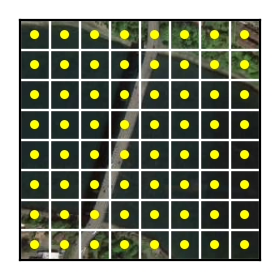}};

\node[draw, trapezium, rotate=-90, fill=gray!15, minimum width=1.5cm, minimum height=0.5cm, trapezium stretches=true] at (4.1,-0.55) (graph_creation) {\makecell{\small2. Image to graph}};

\node at (5.9, -0.55) (global_knn_graph) {\includegraphics[width=2.2cm, height=2.2cm]{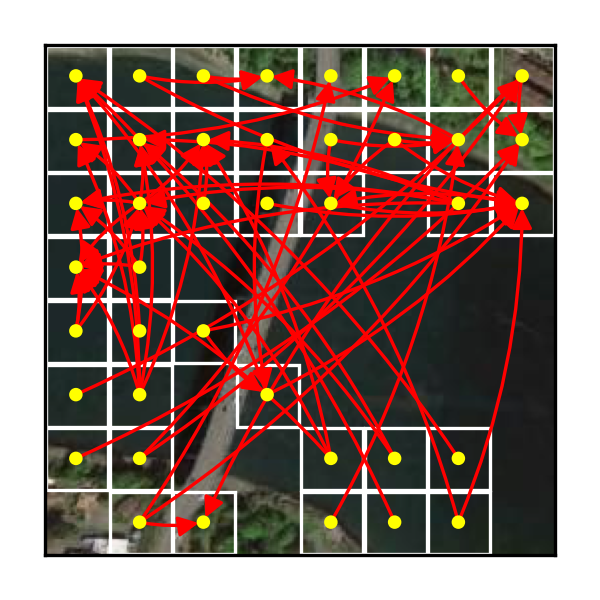}};

\node[draw, trapezium, rotate=-90, fill=gray!15, minimum width=1cm, minimum height=0.5cm, trapezium stretches=true] at (7.75, -0.55) (edge_ranking) {\makecell{
\small{3. Edge ranking}}};

\node at (9.75, -0.55) (sparse_attention) {\includegraphics[width=2.5cm, height=2.5cm]{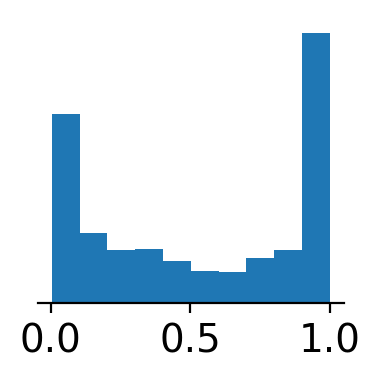}};

\node[draw, trapezium, fill=gray!15, minimum width=1cm, minimum height=0.5cm, rotate=-90, trapezium stretches=true] at (11.8, -0.55) (gnn_bottleneck) {\makecell{4. Global GNN}};

\node at (16.05,-0.55) (explanation) {\includegraphics[width=2.55cm]{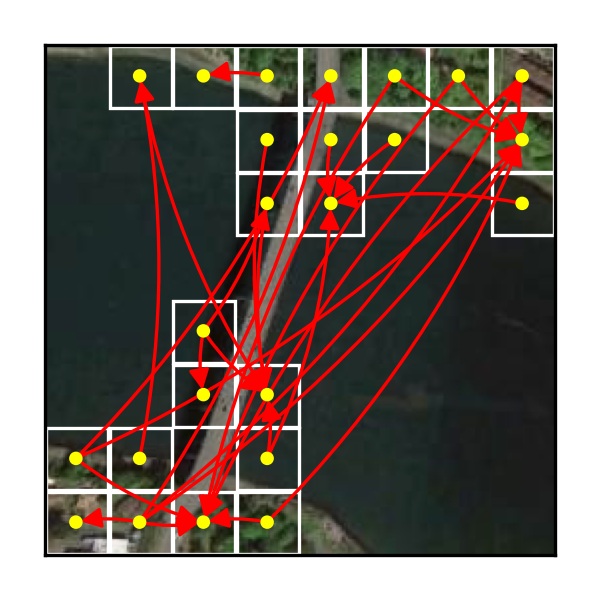}};


\draw [->, ultra thick] (input) -- (bagnet);

\draw [->, ultra thick] (bagnet) -- (feature_map);

\draw [->, ultra thick] (feature_map) -- (graph_creation);

\draw [->, ultra thick] (graph_creation) -- (global_knn_graph);

\draw [->, ultra thick] (global_knn_graph) -- (edge_ranking);

\draw [->, ultra thick] (edge_ranking) -- (sparse_attention);

\draw [->, ultra thick] (sparse_attention) -- (gnn_bottleneck);

\def\graphrep(#1)#2{
\begin{scope}[shift={(#1)}, transform shape]

\foreach \i in {0,1,2,3} {
    \foreach \j in {0,1,2,3} {
        \node[draw=black, fill=yellow, circle, line width=0.5pt, minimum width=0.21cm] at (-0.75 + \j*0.5, 0.75 - \i*0.5) (graphrep\i\j) {};
    }
}

\draw[thick, gray!60] (graphrep00) -- (graphrep01);
\draw[very thick, #2] (graphrep00) -- (graphrep10);
\draw[thick, gray!60] (graphrep10) -- (graphrep11);
\draw[very thick, #2] (graphrep10) -- (graphrep20);
\draw[thick, gray!60] (graphrep10) -- (graphrep21);
\draw[thick, gray!60] (graphrep20) -- (graphrep21);
\draw[very thick, #2] (graphrep20) -- (graphrep31);
\draw[thick, gray!60] (graphrep20) -- (graphrep30);
\draw[thick, gray!60] (graphrep30) -- (graphrep31);

\draw[very thick, #2] (graphrep01) -- (graphrep12);
\draw[thick, gray!60] (graphrep01) -- (graphrep02);
\draw[thick, gray!60] (graphrep01) -- (graphrep11);
\draw[thick, gray!60] (graphrep11) -- (graphrep12);
\draw[very thick, #2] (graphrep11) -- (graphrep21);
\draw[thick, gray!60] (graphrep11) -- (graphrep22);
\draw[very thick, #2] (graphrep21) -- (graphrep22);
\draw[thick, gray!60] (graphrep21) -- (graphrep32);
\draw[very thick, #2] (graphrep21) -- (graphrep31);
\draw[thick, gray!60] (graphrep31) -- (graphrep32);

\draw[very thick, #2] (graphrep02) -- (graphrep13);
\draw[thick, gray!60] (graphrep02) -- (graphrep03);
\draw[very thick, #2] (graphrep02) -- (graphrep12);
\draw[thick, gray!60] (graphrep12) -- (graphrep13);
\draw[very thick, #2] (graphrep12) -- (graphrep22);
\draw[thick, gray!60] (graphrep12) -- (graphrep23);
\draw[thick, gray!60] (graphrep22) -- (graphrep23);
\draw[very thick, #2] (graphrep22) -- (graphrep33);
\draw[thick, gray!60] (graphrep22) -- (graphrep32);
\draw[thick, gray!60] (graphrep32) -- (graphrep33);

\draw[very thick, #2] (graphrep03) -- (graphrep13);
\draw[thick, gray!60] (graphrep13) -- (graphrep23);
\draw[very thick, #2] (graphrep23) -- (graphrep33);
\end{scope}
}

\begin{scope}[canvas is yz plane at x=13.8]
\graphrep(-0.55,0){red} 
\end{scope}
\begin{scope}[canvas is yz plane at x=13.0]
\graphrep(-0.55,0){red}
\end{scope}

\coordinate (gnn_stack) at (12.6, -0.55);
\coordinate (gnn_stack_end) at (14.2, -0.55);

\node at (9.7, 0.9) (attn_text) {\makecell{Sparse edge  \\ attention}};

\node at (13.3, 1.0) (gnn_bottleneck_caption) {\makecell{Global relations}};

\node at (16.05, -1.9) {\textbf{\textsc{Explanation}}};


\draw[blue, thick, rounded corners] (14.8, -2.2) rectangle (17.3,1.4);

\node at (-1.25, 0.75) {Input};

\node at (16.05, 1.0) (outcome_node) {
    \begin{tabular}{c} 
        \small Prediction: \\ 
        \textbf{\textsc{\sffamily Bridge}} 
    \end{tabular}
};

\node at (2.3, 0.9) {\makecell {Non-overlapping\\ Feature Map}};

\node at (6, 0.9) {\makecell{Global k-NN \\  graph}};

\draw[gray, thick, rounded corners, dashed] (9.9, -2.8) rectangle (14.4,-2.3);

\node[gray!60, font=\sffamily\scriptsize\bfseries] at (11., -2.6) {\makecell{$\rule{0.3cm}{0.3cm}$ Low attention}};
\node[red, font=\sffamily\scriptsize\bfseries] at (13.2, -2.6) {\makecell{$\rule{0.3cm}{0.3cm}$ High attention}};

\draw [->, ultra thick] (gnn_bottleneck) -- (gnn_stack);
\draw [->, ultra thick] (gnn_stack_end) -- (explanation);

\draw[black, thick, rounded corners] (7.2, -3) rectangle (14.5, 1.4);
\node[anchor=south west, inner sep=2pt, font=\sffamily\bfseries] at (7.2, -2.76) {GNN Bottleneck};

\end{tikzpicture}}}

\caption{\methodname\ reveals the relevant global relationships for the model prediction with the following steps: (1) A \textbf{non-overlapping local-window encoder} that creates graph nodes with distinct receptive fields in the input image, (2) \textbf{Image to graph} representation with elements of the feature map as nodes and k-NN edges, (3) computing \textbf{sparse edge attention} for ranking the importance of the graph edges, (4) Using the sparse attention to specify the strength of the connections in the \textbf{global GNN} that models the long-range interactions within the image. As an output, our model yields a prediction along with an explanation that depicts the relevant subgraph.}
\label{method_figure}
\end{figure}

We propose i-WiViG (\textbf{I}nterpretable \textbf{W}indow \textbf{Vi}sion \textbf{G}NN), a self-interpretable \gls{vig} model that reveals the relevant subgraph for the model prediction during inference time. It addresses the lack of interpretability in existing \gls{vig} models, which, as shown in Figure \ref{teaser_figure}, struggle to reveal critical spatial interactions. To solve this, our approach (Figure \ref{method_figure}) introduces a novel pyramid \gls{vig} architecture driven  by two key design choices: (1) a \textbf{visual encoder} that produces graph nodes encoding image windows with strictly non-overlapping receptive fields, and (2) a \textbf{graph bottleneck} that estimates sparse edge attention over a global $k$-NN graph. By using this attention to weight the spatial interactions in the image, the model inherently identifies the most relevant subgraph for its final prediction.

\subsection{Non-overlapping Window Encoder}
\label{sec:non_overlapping_encoder}
\begin{wraptable}[11]{r}{0.58\textwidth}
    \vspace{-3\baselineskip}
    \caption{Comparison between the encoder of WiGNet \cite{spadaro2024wignet} and \methodname. We prevent overlap by using a smaller window size, non-overlapping strides, and a smaller number of graph stages.}
    \centering
    \setlength{\tabcolsep}{5pt} 
    \begin{tabular}{lcc} \toprule
         \textbf{Model} & WiGNet  & \methodname\ (ours) \\ \midrule
         Window size & 8 & 4 \\
         Kernel size/stride &  3 / 2 & 2 / 2 \\
         \makecell[l]{GNN Bottleneck\\ Kernel size/stride} &  3 / 2 & 4 / 4 \\
         Graph Stages & 3 & 2 \\ \bottomrule
    \end{tabular}
    \label{tab:comparison_vig_family}
\end{wraptable}
The existing \gls{vig} approaches produce large and overlapping receptive fields. This limits the understanding of how the graph exchanges information between different objects in the image to derive the prediction. As illustrated in Figure \ref{teaser_figure}, this behavior also holds for \gls{wignet}, even though it constrains the graph processing to local image windows. In \gls{wignet}, this is caused by:
\begin{itemize}
    \item Using a default window size of 8, which means that the local window graph operates on a neighbourhood of $8 \times 8$ pixels. 
    \item The downsampling operation, which is similar to the other \gls{vig} approaches, still utilizes a kernel size of 3 and a stride of 2. This introduces overlap between nodes within the same window and increases the receptive field and the overlap of the nodes located on the edges of the windows.
\end{itemize}

To produce graph nodes with non-overlapping receptive fields, as illustrated in Table \ref{tab:comparison_vig_family}, we use a window size of $4$, kernel size and stride of 2 in the earlier downsampling layers, and kernel size and stride of 4 in the latest encoder layer. This way, we achieve the same final encoder feature map size of $8 \times 8$ as in \gls{wignet}; however, in our approach, each node in the feature map encodes a distinct receptive field. An example of the non-overlapping feature map by our encoder is shown in Figure \ref{method_figure} and the detailed architecture of \methodname\ is provided in the Appendix, Table \ref{tab:iWiViG_architecture}.

\subsection{Graph Bottleneck with Sparse Edge Attention}
\label{sec:method_edge_ranking}
\begin{wrapfigure}[4]{r}{0.18\textwidth} 
    \vspace{-6\baselineskip}
    \centering
    \includegraphics[trim={0cm 0cm 0cm 0cm}, clip, width=\linewidth]{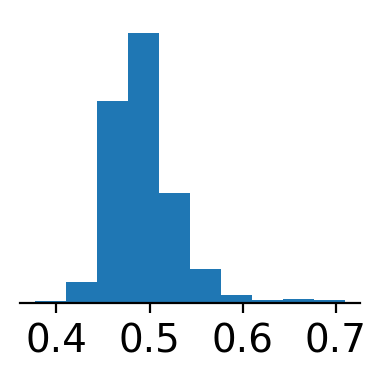}
    \caption{GSAT edge attention}
    \label{fig:gsat_edge_attention}
\end{wrapfigure}
The \gls{vig} models typically assign uniform weights to all connections in the graph. As such, they lack a built-in mechanism to identify the relevant spatial interactions in the image. Although inherently interpretable \gls{gnn} approaches can be used for this, they are usually designed to operate on data with a natural graph representation. For instance, the loss objective from the \gls{gsat} approach in Equation \ref{eq:gsat} encourages densely centered attention distribution, as the one visualized in Figure \ref{fig:gsat_edge_attention} for the SUN397 dataset, where edge attention is clustered around $0.5$ with a low variance. 
While this approach may work well for graphs, such as molecules or motifs \cite{miao2022interpretable}, where the relevant structure can be captured in a small subgraph described by a few edges with higher attention value than the mean, applying it to image graphs might be problematic, as it does not disregard the redundant and noisy edges that often appear in image graphs \cite{han2023vision}.

To overcome this problem and only assign high attention to the edges that matter for the prediction, we employ a \gls{gnn} bottleneck that uses the above \gls{gsat} mechanism to estimate the edge attention, but instead enforces sparse attention distribution $\alpha$ over the graph edges with the following loss function: 
\begin{equation}
\label{eq:sparse_attention_loss}
L_{\text{sparsity}}(\alpha) = \lambda_{\text{G}} \sum_{i=1}^{|E|} \alpha_i (1-\alpha_i) + \lambda_{\text{B}} \sum_{i=1}^{|E|} (\alpha_i - B)^2,
\end{equation}
where the first term constitutes Gini impurity, which enforces a bimodal distribution of the edge attention around 0 and 1, and the second term helps control the level of sparsity in the graph, with the sparsity hyperparameter $B$, and $\lambda_G, \lambda_B$ are weights. The sparsity loss is similar to the one employed in the PGExplainer method \cite{NEURIPS2020_e37b08dd}, with the difference that we apply it during model training, instead of post-hoc. 
With this, the loss term of \methodname\ consists of: 
\begin{equation}\mathcal{L} =  \mathcal{L}_{\text{task}} + \mathcal{L}_{\text{sparsity}}
\end{equation}
where $\mathcal{L}_{\text{task}}$ corresponds to the cross-entropy loss for classification tasks and mean squared error for regression.



\section{Experimental Setup}
\label{sec:experimental_setup}
\paragraph{Datasets}
We evaluate our model on the tasks of scene classification and image regression from natural and remote sensing imagery. For the first classification task, we utilized the SUN397 dataset \cite{5539970}, which poses the problem of categorizing a natural image into one of 397 distinct scenes. We used the default splits: Training\_01 for training, Testing\_01 for validation, and Testing\_02 for testing. Each of those splits contains 50 images per class. For the second task, we used the NWPU-RESISC45 dataset, where the goal is to classify $31,500$ aerial images into $45$ scene classes \cite{cheng2017remote}. We used the dataset splits provided in \cite{neumann2019domain}, which assign 60\% of the instances to the training set, 20\% to the validation set, and the remaining 20\% to the test set, respectively. Finally, for the regression task, we used the Liveability dataset, where instances associate aerial imagery covering a neighborhood area in the Netherlands with its liveability score \cite{levering2023predicting}. The liveability is a composite score derived from statistics about the physical environment, buildings, amenities, population, and safety. This dataset includes $51,781$ labeled images and covers 13 cities in the Netherlands. For our experiments, we used the geographically stratified splits provided by the authors, which test the model generalization on unseen cities during model training.

\paragraph{Benchmark Models}
\paragraph{Model Training}
We train all models for 300 epochs using a Cosine Annealing schedule with an initial learning rate of $0.001$ and a weight decay of $0.05$. For the RESISC45 dataset, we extend the training of \methodname\ to 500 epochs to ensure optimal convergence. 
During training, we evaluated the model's performance every 30 epochs on the validation set, and for testing, we selected the model with the best validation performance. For our sparse attention loss in Equation \ref{eq:sparse_attention_loss}, we set the sparsity budget $B$ to $0.7$, $\lambda_{\text{B}}$ to $0.1$, and $\lambda_{\text{G}}$ to $0.01$ for the classification tasks and to $0.001$ for the regression task. 
Further details about the training procedure and image augmentations are provided in Appendix, Section \ref{sec:appendix_training_procedure}. 

\section{Results}

\subsection{Model Performance}



\begin{table}[h]
\caption{\textbf{Model performance on the test sets.}}
\label{tab:model_performance_results}
\centering
\scriptsize
\begin{tabular}{ll S[table-format=1.2] S[table-format=1.2] S[table-format=1.2] S[table-format=1.2]}
\toprule
& & \multicolumn{1}{c}{\textbf{SUN397}} & \multicolumn{1}{c}{\textbf{RESISC45}} & \multicolumn{2}{c}{\textbf{Liveability}} \\
\cmidrule(lr){3-3} \cmidrule(lr){4-4} \cmidrule(lr){5-6} 
\textbf{Approach} & \textbf{Model} & {Acc} & {Acc} & {$R^2$} & {$\tau$} \\ 
\midrule
\multirow{2}{*}{CNN} & BagNet & 0.60 & 0.91 & 0.42 & 0.51 \\
                     & ResNet-18 & \bfseries 0.62 & 0.93 & 0.38 & 0.46 \\
\midrule
\multirow{3}{*}{ViG} & Isotropic ViG & 0.60 & 0.94 & 0.43 & 0.49 \\
                                 & Pyramid ViG   & 0.59 & 0.94 & \bfseries 0.48 & \bfseries 0.52 \\
                                 & WiGNet & 0.61 & \bfseries 0.95 & 0.42 & \bfseries 0.52 \\
\midrule
\textbf{Ours} & \textbf{\methodname} & 0.58 & 0.92 & 0.47 & 0.50 \\
\bottomrule
\end{tabular}
\end{table}

\begin{figure*}[t!]
    \centering    
    \subfloat[Class "Waterfall" (SUN397)]{{\includegraphics[width=0.33\textwidth]{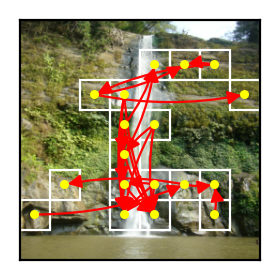}}\label{fig:qualitative_sun397}}
    \subfloat[Class "Airplane" (RESISC45)]{
        {\includegraphics[width=0.33\textwidth]{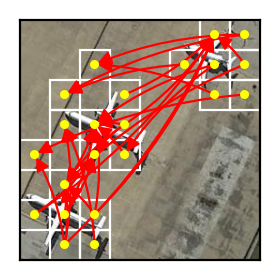} }\label{fig:qualitative_resisc45}}
    \subfloat["High Liveability"]{\vspace{0cm}{\includegraphics[width=0.33\textwidth]{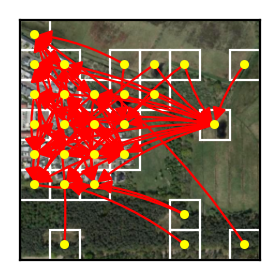}}\label{fig:qualitative_liveability}}
    \caption{Qualitative explanations for examples from the class \textit{Waterfall} from the SUN397 dataset  (left), class \textit{Airplane} from the RESISC45 dataset (middle), and the Liveability dataset (right). The visualized subgraphs contain the edges with the top 5\% of attention scores.}
    \label{fig:qualitative_xai}
\end{figure*}

Table \ref{tab:model_performance_results} compares the performance of our \methodname\ model with the different CNN and \gls{vig} benchmarks. Overall, despite the constraint caused by introducing the interpretable bottleneck, our model exhibits comparable performance to the baselines. In detail, the third column indicates that for the SUN397 dataset, \methodname\ yields an accuracy of $0.58$, which is slightly worse but still comparable to the \gls{cnn} and \gls{vig} benchmarks, whose accuracy ranges from $0.59$ to $0.62$. 
Next, the results for the RESISC45 scene classification dataset in the fourth column reveal that our self-interpretable \methodname\ approach achieves an accuracy of $0.92$. As such, it yields competitive performance compared to the \gls{vig} and \gls{cnn} benchmarks, and slightly surpasses the performance of the BagNet model. Finally, regarding the Liveability dataset, the last two columns of Table \ref{tab:model_performance_results} demonstrate that our approach improves the $R^2$ score by approximately $0.1$ over the \gls{cnn} baseline and $0.05$ over the Bagnet and the \acrshort{vig} baselines. At the same time, it exhibits comparable performance to the Pyramid \gls{vig}. Furthermore, while the Kendall $\tau$ scores in the last column show that the models produce a similar ranking of the liveability scores, the higher $R^2$ value indicates that \methodname\ is more precise in the fine-grained prediction of the actual liveability values. These results suggest that even on datasets exhibiting strong texture bias, such as SUN397 and RESISC45, our method can achieve competitive performance by enforcing interpretable reasoning based on spatial interactions, rather than local texture.

\subsection{Qualitative Analysis of the Relevant Subgraphs}

\methodname\ learns sparse attention across edges connecting nodes with distinct, non-overlapping receptive fields in the image. This mechanism inherently explains the model's behavior by highlighting the key spatial interactions for its predictions. Consequently, it offers a significant interpretability advantage over existing \gls{vig} methods, which do not preserve spatial relationships between graph nodes and their positions in the image. Figure \ref{fig:qualitative_xai} presents qualitative examples of our model's explanations across the datasets used in our experiments. In detail,
the left Figure \ref{fig:qualitative_sun397} illustrates that the model leverages a sparse subgraph to associate a waterfall example from the SUN397 dataset with its surrounding features, including the cavities at its base. Further, the middle
Figure \ref{fig:qualitative_resisc45} demonstrates that, for the airplane example from the RESISC45 dataset, our model identifies a sparse subgraph capturing long-range dependencies between multiple airplanes, while also modelling local connections within individual airplanes -- such as the edges linking the tail to the wings. Finally, the right
Figure \ref{fig:qualitative_liveability} presents an explanation example for the Liveability regression task. The most important edges cluster houses into a connected component that links them to nearby vegetation and trees, thus demonstrating the importance of the connections within the residential area and the nearby natural environment for predicting high liveability scores. Overall, these examples demonstrate that our approach effectively uncovers both critical local relationships and long-range dependencies underlying the model’s predictions. Further qualitative explanation examples are provided in the Appendix, Section \ref{sec:appendix_qualitative_explanation_evaluation}.

\subsection{Quantitative Explanation Evaluation}
\begin{figure*}[t!]
    \centering
    \subfloat[SUN 397]{
    \includegraphics[width=0.33\textwidth]{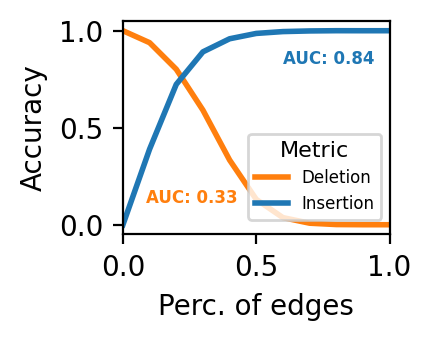}
    \label{fig:exp_eval_sun397}}
     \subfloat[RESISC45]{
    \includegraphics[width=0.33\textwidth]{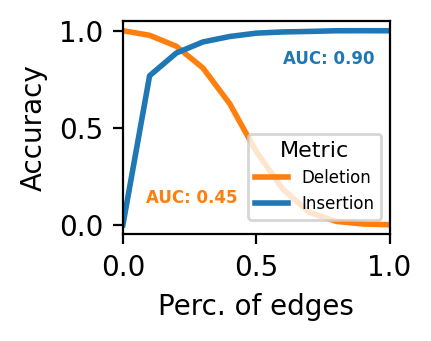}
    \label{fig:exp_eval_resisc45}}
     \subfloat[Liveability]{
    \includegraphics[width=0.33\textwidth]{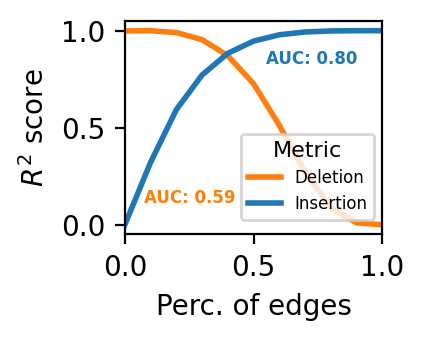}
    \label{fig:exp_eval_liveability}}
    \caption{Quantitative evaluation of the \methodname\ subgraph explanations with the insertion (blue curve) and deletion (orange curve) metrics. The Area under the Curve (AUC) scores are min-max normalized.
    } 
    \label{fig:iwivig_quantitative_eval}
\end{figure*}
\label{sec:quantiative_evaluation}
To assess whether the explanation subgraphs produced by \methodname\ faithfully explain the model workings, we quantitatively evaluate the \textit{insertion} and \textit{deletion} explanation metrics presented in \cite{Petsiuk2018rise}. The insertion metric measures the increase in model performance as the edges with high attribution are sequentially introduced to an empty bottleneck graph. Conversely, the deletion metric measures performance decay as the most important connections are sequentially excluded from the graph. Faithful explanations are characterized by a high Area Under the Curve (AUC) for the insertion metric and a low AUC for the deletion model. In the context of our model, these results would indicate that the identified subgraphs reveal the critical spatial interactions in the image for the model's predictions. Figure \ref{fig:iwivig_quantitative_eval} presents the results of this experiment, illustrating that across all three tasks, our sparse edge attention strongly correlates with the prediction performance. 
Specifically, the blue insertion curves indicate that introducing the edges with high attention leads to a sharp improvement in model performance. Complementing these results, the insertion curves show substantially higher AUC than the orange deletion curves, with improvements of a factor of around 2.5, 2, and 1.35 on the SUN397 and RESISC45 datasets, and the Liveability dataset, respectively.   Finally, we observe that the induced sparsity constraint in our loss function effectively prunes less informative edges, as Figure \ref{fig:iwivig_quantitative_eval} shows that the performance remains nearly stagnant when adding the bottom 60 \% of edges on all three datasets.

\subsection{Comparison with \gls{gnn} explanation methods}

\begin{table}[t]
\caption{Comparison of the explanation properties of \methodname\ with post-hoc GNNExplainer method, and the inherently interpretable \gls{gsat} approach. $AUC_{ins}$ and $AUC_{del}$ denote the Area Under the Curve for insertion and deletion metrics, respectively. Bold values indicate the best performance for each explanation metric. While the different models yield similar performance, \methodname\ produces the most faithful explanations.
\label{tab:explanation_quality_comparison}}

\centering
\scriptsize
\setlength{\tabcolsep}{0.5pt}
\begin{tabular*}{\textwidth}{@{\extracolsep{\fill}} l c cc c cc c cc}
\toprule
 & \multicolumn{3}{c}{\textbf{SUN397}} & \multicolumn{3}{c}{\textbf{RESISC45}} & \multicolumn{3}{c}{\textbf{Liveability}} \\
 \cmidrule(lr){2-4} \cmidrule(lr){5-7} \cmidrule(lr){8-10}
\textbf{Approach} & Acc. & \textbf{$AUC_{ins} \uparrow$} & \textbf{$AUC_{del} \downarrow$} & Acc. & \textbf{$AUC_{ins} \uparrow$} & \textbf{$AUC_{del} \downarrow$} & $R^2$ & \textbf{$AUC_{ins} \uparrow$} & \textbf{$AUC_{del} \downarrow$} \\
\midrule
GNNExplainer & $0.55$ & $0.55$ & $0.36$ & $0.91$ & $0.74$ & $0.52$ & $0.47$ & $0.75$ & $0.73$ \\
GSAT & $0.58$ & $0.74$ & $0.7$ & $0.93$ & $0.81$ & $0.81$ & $0.47$ & \textbf{0.82} & 0.83 \\
\midrule
\textbf{\methodname} & $0.58$ & \textbf{0.84} & \textbf{0.33} & $0.92$ & \textbf{0.90} & \textbf{0.45} & $0.47$ & 0.80 & \textbf{0.59} \\
\bottomrule
\end{tabular*}

\end{table}

Having demonstrated that the estimated sparse edge attention in \methodname\ faithfully explains its predictions, we further compare the properties of our explanations with those of the post-hoc GNNExplainer \cite{ying2019gnnexplainer} approach, and the inherently interpretable \gls{gsat} \cite{miao2022interpretable} method. For this, we train a new model that uses the same non-overlapping window encoder as \methodname. For GNNExplainer, we disabled the sparsity loss. This effectively assigns a uniform weight to all edges during message passing. After the model training, we apply GNNExplainer to estimate the relevant edges of the instances in the test set. Regarding \gls{gsat}, we employ its loss function in Equation \ref{eq:gsat} to estimate the edge attention during training by setting the $r$ hyperparameter to 0.5.

The results of this experiment are displayed in Table \ref{tab:explanation_quality_comparison} and show that even though the different models produce similar accuracy, \methodname\ substantially outperforms GNNExplainer and \gls{gsat} in terms of explanation quality. Specifically, our approach achieves by far the largest AUC margin between the insertion and deletion metrics across all datasets. As shown in the second row, \gls{gsat} produces nearly identical AUC scores for the insertion and deletion metrics, which indicates that it struggles to distinguish between class discriminative and redundant connections in the image graph. Furthermore, while the results for GNNExplainer in the first row show that its explanations produce more discriminative curves than \gls{gsat}, the resulting gap remains narrower than that of our proposed method (e.g., $0.45$ of our method vs $0.24$ of GNNExplainer on the RESISC45 dataset). The suboptimal performance of GNNExplainer can be attributed to its post-hoc nature, which is often shown to yield unreliable explanations of the underlying model's workings \cite{adebayo2018sanity, miao2022interpretable}. On the other hand, the \gls{gsat} approach produces a narrow edge attention distribution with a small standard deviation that is highly centered around the $r$ hyperparameter in Equation \ref{eq:gsat} (Appendix, Table \ref{tab:explanation_quality_comparison_compact}). 
Our results show that this inductive bias is ineffective for distinguishing important connections from unimportant ones in complex vision problems, where graphs encode noisy, redundant interactions among different image parts.

\begin{figure*}[t!]
    \centering    
    \subfloat[\methodname]{{\includegraphics[width=0.33\textwidth]{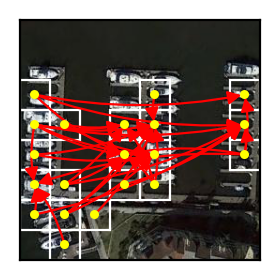}}\label{fig:qualitative_iwivig_harbor}}
    \subfloat[GNNExplainer]{
        {\includegraphics[width=0.33\textwidth]{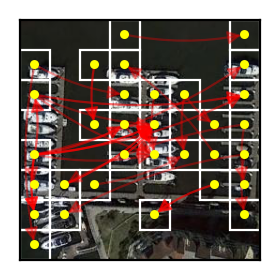} }\label{fig:qualitative_gnnexplainer}}
    \subfloat[GSAT]{\vspace{0cm}{\includegraphics[width=0.33\textwidth]{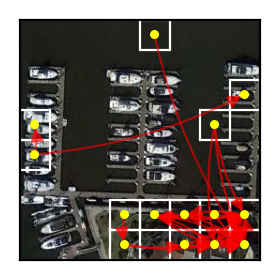}}\label{fig:qualitative_gsat}}
    \caption{Qualitative explanations for an example of the Harbor class from the RESISC45 dataset from our approach (left) with the GNNExplainer (middle), and \gls{gsat} methods. The visualized subgraphs contain the edges with the top 5\% of attention scores.}
    \label{fig:qualitative_xai_methods_comparison}
\end{figure*}

Further, we qualitatively compare the explanations of \methodname\ against those of GNNExplainer and \gls{gsat} in Figure \ref{fig:qualitative_xai_methods_comparison} and Figures \ref{fig:qualitative_xai_resisc45}, \ref{fig:qualitative_xai_sun397}, and \ref{fig:qualitative_xai_liveability} in the Appendix. Overall, these visualizations corroborate the quantitative advantages reported in Table \ref{tab:explanation_quality_comparison}. While GNNExplainer suffers from noisy artifacts and \gls{gsat} focuses on local spurious patterns, driven by the sparsity loss, \methodname\ produces more compact subgraphs that highlight long-range, semantically meaningful interactions. For example, the explanations for the Harbor instance in Figure \ref{fig:qualitative_xai_methods_comparison} show that \methodname\ yields a sparser explanation compared to GNNExplainer, as it highlights interactions among ships while GNNExplainer also assigns relevance to connections among water and land. Second, our method reveals more semantically related associations than the \gls{gsat} explanation, which identifies a localized cluster that mostly describes interactions on land, largely ignoring the ships. 

\subsubsection{Impact of Non-overlapping Receptive Fields}

Lastly, we evaluate how enforcing non-overlapping receptive fields affects model performance and the quality of explanations. More specifically, we train new models where we only replace the non-overlapping window encoder in \methodname\ with the \gls{pvig} and \gls{wignet} encoders, which yield larger and overlapping node receptive fields. We also consider a third encoder based on \gls{wignet}, namely \gls{wignet} (Conv $\times 2$), which reduces the receptive field and overlap of \gls{wignet} by using a convolution kernel size and stride of 2 for the downsampling operation. After model training, we measure its performance
and the gap in the AUC ($\Delta AUC$) between the insertion and deletion metrics.  
The results of this experiment are presented in Table \ref{tab:local_windows_impact}, which demonstrates that reducing the size of the node receptive field and its overlap with other nodes substantially improves the quality of subgraph explanations and can also enhance the model performance. Further, while the \gls{wignet} (Conv $\times 2$) encoder that produces overlapping nodes of size 4096px yields better explanation quality on Resisc45 and Liveability datasets than our proposed method, as shown in Figure \ref{fig:wignn_conv2_windows} in the appendix, due to the node overlap, it produces explanations that are more difficult to interpret as they struggle to reveal the relevant interactions among fine-grained objects and contain self-loops. Therefore, the \methodname\ encoder, which produces disjoint graph nodes with a receptive field of 1024 pixels, offers the optimal trade-off between model accuracy and intuitive explanations with high faithfulness across datasets.

\begin{table*}[t!]
\caption{Impact of the size of the encoder's receptive field and node overlap on the model performance and explanation quality. The size of the receptive field is calculated based on the input image size of $2^8$ $\times$ $2^8$ px. The overlap column indicates whether graph nodes share the same receptive field in the image. Each encoder outputs a feature map of $8 \times 8$.}
\label{tab:local_windows_impact}
\centering
\scriptsize
\begin{tabular}{lllllllll}
\toprule
 &  &  &  \multicolumn{2}{c}{\textbf{SUN397}} & \multicolumn{2}{c}{\textbf{RESISC45}} & \multicolumn{2}{c}{\textbf{Liveability}} \\
\cmidrule(lr){4-5}
\cmidrule(lr){6-7}
\cmidrule(lr){8-9}
\textbf{Encoder} & \textbf{Size} &  \textbf{Overlap} & \textbf{Acc.} & \textbf{ $\Delta AUC$} & \textbf{Acc.} & \textbf{ $\Delta AUC$} & \textbf{$R^2$} & \textbf{$\Delta AUC$} \\
\midrule

PViG & $2^{16}$ & \cmark & $0.56$ & $0.28$  & $0.93$ & $-0.09$ & $0.44$ & $0.23$ \\

WiGNet &  $2^{14}$ & \cmark & $0.55$ & $0.42$ & $0.94$ & $0.38$ & $0.41$ & $0.21$   \\

WiGNet (Conv $\times 2$) & $2^{12}$ & \cmark & $0.55$ & $0.4$ & $0.92$ & $0.49$ & $0.46$ & $0.29 $ \\ 
\midrule

\methodname & $2^{10}$ & \xmark & $0.58$ & $0.51$ & $0.92$ & $0.45$ & $0.47$ & $0.21$ \\
\bottomrule
\end{tabular}
\end{table*}

\section{Conclusion}
This work addresses the lack of interpretability in vision graph neural networks by introducing \textbf{\methodname}, a novel, inherently explainable \gls{vig} approach, capable of automatically identifying sparse subgraphs containing the relevant long-range dependencies for the model prediction. We achieve this by introducing 1) a visual encoder that maintains the spatial consistency of the graph nodes to non-overlapping local image windows, and 2) a graph bottleneck with an interpretable long-range dependency modeling between the local windows based on sparse edge attention. 
We perform extensive qualitative and quantitative evaluations of our approach on the tasks of scene classification and image regression from natural and remote sensing imagery. The qualitative analysis indicates that our approach produces sparse subgraphs that intuitively describe the model decision mechanism. In detail, Figures \ref{fig:qualitative_xai} and \ref{fig:qualitative_xai_methods_comparison} 
reveal that our method identifies subgraphs  describing semantic interactions closely related to the target outcome that are sparser and contain less noisy artifacts compared to the benchmark \gls{gnn} explainability methods. These insights are further validated by the quantitative explanation evaluation, which shows that our method produces explanations (Figure \ref{fig:iwivig_quantitative_eval}) with substantially higher fidelity than those from benchmark \gls{gnn} explanation methods, underpining the importance of the sparse attention mechanism and the non-overlapping receptive fields (Table \ref{tab:explanation_quality_comparison}).  Finally, Table \ref{tab:model_performance_results} shows that our method achieves competitive predictive performance compared to state-of-the-art black-box counterparts, even on datasets with strong texture bias. This indicates that enforcing interpretable model reasoning in \glspl{vig} does not sacrifice predictive performance.

\section*{Acknowledgements}
The work began during the research visit of Ivica Obadic to the University of Lancaster, supported by TAILOR, a project funded by the EU Horizon 2020 research and innovation programme under Grant Agreement No. 952215. Further, the work of Ivica Obadic, Adrian Höhl, and Xiaoxiang Zhu has been supported by the Munich Center for Machine Learning and by the ML4Earth project of the German Federal Ministry for Economic Affairs and Energy under grant number 50EE2201C. The work of Dmitry Kangin and Plamen Angelov was supported by the European Union’s
ELSA – European Lighthouse on Secure and Safe AI, Horizon Europe, Grant Agreement No. 101070617.

%
%
\bibliographystyle{splncs04}
\bibliography{main}

\clearpage
\setcounter{page}{1}

\appendix

\section{\methodname\ Architecture}
\label{sec:appendix_architecture}

Table \ref{tab:iWiViG_architecture} presents in detail the architecture of our proposed \methodname\ model and its hyperparameters. Similar to the other pyramid \gls{vig} models, it alternates between graph processing stages and \gls{cnn} downsampling layers. It consists of three \gls{gnn} stages, with the last stage followed by a pooling layer and an MLP to derive the prediction. As stated in Section \ref{sec:method}, \methodname\ is based on two design objectives:
\begin{enumerate}
    \item \textbf{Graph nodes with non-overlapping receptive fields:} This objective is fulfilled by the two initial stages in the model and the subsequent downsampling operations. The stages are based on the \gls{wignet} local graph processing blocks with a window size $W$ of 4. While other VIG models downsample by halve the feature map using a kernel of $3$ and a stride of $2$ (thereby introducing overlap), our approach uses CNN layers with an equal kernel size and stride for downsampling to prevent overlap. Particularly, we use a kernel size and a stride of $2$ for downsampling the feature map after the first graph stage. For a local image window of $4 \times 4$ pixels, this operation downsamples it to a window of $2 \times 2$, with the remaining pixels sharing the same receptive field. After the second graph stage, to prevent any overlap between the nodes, we use a kernel size and a stride of $4$ that reduces the feature map size by 4. This way, it produces a feature map with an output size of $\frac{H}{32}$ $\times$ $ \frac{W}{32}$, where each pixel encodes a distinct receptive window in the input image. Crucially, this design allows our architecture to have one stage fewer than other VIG models, which typically use four stages where the first three transform the input image to the same $\frac{H}{32} \times \frac{W}{32}$ feature map size. This results in a decrease in parameters from $13.3$M when using the default \gls{wignet} encoder to $ 12.2$M with our encoder, which produces nodes with smaller and distinct receptive fields.
    \item \textbf{\gls{gnn} bottleneck to identify the relevant global connections:} 
    In the third stage, we construct a global $k$-nearest neighbours graph between the local image windows and employ an interpretable \gls{gnn} bottleneck that first estimates the edge attention, encouraging sparse attention with the sparsity objective in Equation \ref{eq:sparse_attention_loss}. Afterwards, the edge attention is used to weight the importance of global connections between the local image windows when deriving the model prediction. As a consequence of this reasoning process, the edge attention can be used to explain the relevant subgraph for the prediction. Similar to the standard grapher layers in the \gls{vig} models, in this stage, we also perform the operations in the following order on a graph node:
    \begin{enumerate}
        \item Linear transformation of the node embeddings
        \item GIN graph convolution
        \item Residual operation
        \item FFN layer processing
    \end{enumerate}
\end{enumerate}

\begin{table}[ht!]
\caption{
\textbf{Detailed configuration of the \methodname\ model.} D: graph nodes feature dimension, E: hidden dimension ratio in FFN, k: number of neighbors in the \gls{gnn}, W: window size, H × W: input image size. The \gls{wignet} blocks use the Max-Relative Graph Conv   operation \cite{li2019deepgcns} while the \gls{gsat} block uses the GIN Conv operation  \cite{xu2018powerful} for graph processing.}
\label{tab:iWiViG_architecture}
\centering
\scriptsize
\begin{tabular}{llllllll}
\toprule
\textbf{Stage} & \textbf{Output Size} & \textbf{Hyperparameters} 
\\ \midrule

Stem  & $\frac{H}{4} \times \frac{W}{4}$ &  Conv $\times 2$ \\
\midrule

Stage 1 \\ \gls{wignet} block & $\frac{H}{4} \times \frac{W}{4}$  & $\begin{bmatrix}
  D=48 \\
  E=4 \\
  k=9 \\
  W=4
\end{bmatrix} \times 2 $  \\ 
\midrule

Downsample & $\frac{H}{8} \times \frac{W}{8}$  & Conv $\times 2$  \\ 
\midrule

Stage 2 \\ \gls{wignet} block & $\frac{H}{8} \times \frac{W}{8}$  & $\begin{bmatrix}
  D=192 \\
  E=4 \\
  k=9 \\
  W=4
\end{bmatrix} \times 4$  \\ 
\midrule

Downsample & $\frac{H}{32} \times \frac{W}{32}$  & Conv $\times 4$  \\ 
\midrule

Stage 3 \\ Interpretable \\ \gls{gnn} 
bottleneck & $\frac{H}{32} \times \frac{W}{32}$  & $\begin{bmatrix}
  D=384 \\
  E=4 \\
  k=9 \\
\end{bmatrix} \times 3$  \\ 
\midrule
Head & $1 \times 1$ & Pooling \& MLP \\ 
\midrule
Parameters (M) &  & 12.2M \\ 
\bottomrule
\end{tabular}
\end{table}

\section{\methodname\ Training Procedure}
\label{sec:appendix_training_procedure} 

For model training, we performed image transformations in the following order:
\begin{enumerate}
    \item Resizing the images to a size of 256 x 256
    \item Random Augmentation 
    \item Random Erasure 
    \item Min-max image normalization
\end{enumerate}
Furthermore, for the image classification tasks, we employed the Cutmix \cite{yun2019cutmix} and Mixup \cite{zhang2017mixup} augmentations during training.



\section{Edge Attention Distribution}
\label{sec:appendix_edge_attention_distribution}

In Table \ref{tab:explanation_quality_comparison_compact}, we relate the distribution of the edge attention of the different explanation methods with the model accuracy and the explanation quality. While the different models display similar performance, the sparsity loss in \methodname\ (Eq. \ref{eq:sparse_attention_loss} in Section \ref{sec:method_edge_ranking}) produces the highest standard deviation in edge attention. Consequently, as shown in the $\Delta AUC$ column, the \methodname\ yields explanations with substantially higher faithfulness than GNNExplainer and GSAT.

\begin{table}[t]
\caption{Comparing the average edge attention ($\bar{\alpha}$), standard deviation, and the difference between the AUC of the insertion and deletion ($\Delta AUC$) between our approach and the benchmark GNN explanation methods.}
\label{tab:explanation_quality_comparison_compact}

\centering
\scriptsize
\setlength{\tabcolsep}{1.2pt} 

\begin{tabular*}{\textwidth}{@{\extracolsep{\fill}} l ccc ccc ccc}
\toprule
 & \multicolumn{3}{c}{\textbf{SUN397}} & \multicolumn{3}{c}{\textbf{RESISC45}} & \multicolumn{3}{c}{\textbf{Liveability}} \\
 \cmidrule(lr){2-4} \cmidrule(lr){5-7} \cmidrule(lr){8-10}
\textbf{Approach} & Acc. & $\bar{\alpha}$ & $\Delta AUC \uparrow$ & Acc. & $\bar{\alpha}$ & $\Delta AUC \uparrow$ & $R^2$ & $\bar{\alpha}$ & $\Delta AUC \uparrow$ \\
\midrule
GNNExplainer & $0.55$ & $0.63 \pm .09$ & $0.21$ & $0.93$ & $0.38 \pm .13$ & $0.22$ & $0.47$ & $0.27 \pm .03$ & $0.02$ \\
GSAT & $0.58$ & $0.50 \pm .03$ & $0.04$ & $0.91$ & $0.49 \pm .10$ & 0.0 & $0.47$ & $0.50 \pm .00$ & $-0.1$ \\
\midrule
\textbf{\methodname} & $0.58$ & \textbf{0.40 $\pm$ .35} & \textbf{0.51} & $0.92$ & \textbf{0.49 $\pm$ .41} & \textbf{0.45} & $0.47$ & \textbf{0.70 $\pm$ .39} & \textbf{0.21} \\
\bottomrule
\end{tabular*}

\vspace{-10pt}
\end{table}

\section{Qualitative Explanation Evaluation}

\subsection{Comparison with \gls{gnn} explanation methods}
\label{sec:appendix_qualitative_explanation_evaluation}
\begin{figure*}[t!]
    \centering
   \makebox[\textwidth]{\textbf{\methodname}}
    \vspace{2mm} 
    \subfloat[Class "Railway Station"]{{\includegraphics[width=0.33\textwidth]{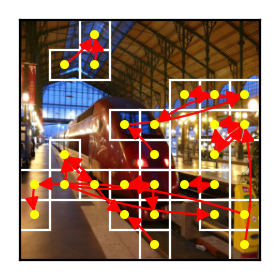}}\label{fig:qualitative_sun397_railway}}
    \subfloat[Class "Village"]{
        {\includegraphics[width=0.33\textwidth]{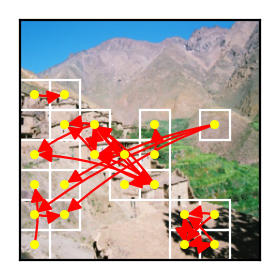} }\label{fig:qualitative_sun397_village}}
    \subfloat[Class "Volleybal court"]{\vspace{0cm}{\includegraphics[width=0.33\textwidth]{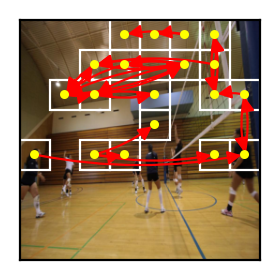}}\label{fig:qualitative_sun_volleybal_court}}

    \makebox[\textwidth]{\textbf{GNNExplainer}}
    \vspace{-3mm} 
    
    \subfloat[Class "Railway Station"]{{\includegraphics[width=0.33\textwidth]{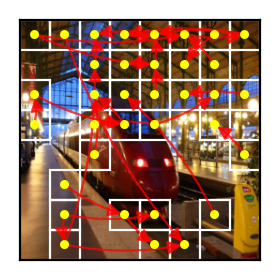}}\label{fig:qualitative_sun397_gnn_explainer_railway}}
    \subfloat[Class "Village"]{
        {\includegraphics[width=0.33\textwidth]{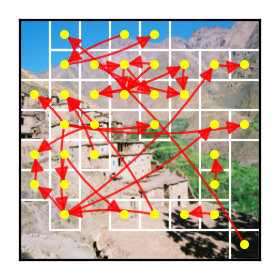} }\label{fig:qualitative_sun397_gnnexplainer_village}}
    \subfloat[Class "Volleybal court"]{\vspace{0cm}{\includegraphics[width=0.33\textwidth]{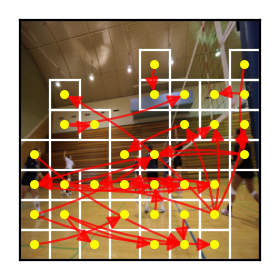}}\label{fig:qualitative_sun_gnnexplainer_volleybal_court}}

    \makebox[\textwidth]{\textbf{GSAT}}
    \vspace{-3mm} 
    
    \subfloat[Class "Railway Station"]{{\includegraphics[width=0.33\textwidth]{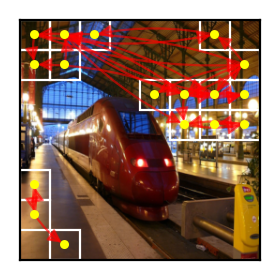}}\label{fig:qualitative_sun397_gsat_railway}}
    \subfloat[Class "Village"]{
        {\includegraphics[width=0.33\textwidth]{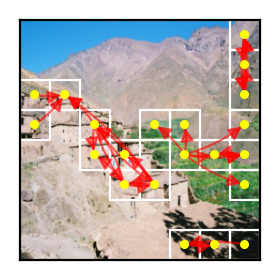} }\label{fig:qualitative_sun397_gsat_village}}
    \subfloat[Class "Volleybal court"]{\vspace{0cm}{\includegraphics[width=0.33\textwidth]{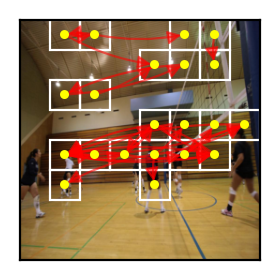}}\label{fig:qualitative_sun_gsat_volleybal_court}}
    
    \caption{\textbf{Qualitative comparison of the explanations of our \methodname\ model (top row) against those from GNNExplainer (middle row) and \gls{gsat} (bottom row) on the SUN397 dataset for classes \textit{Railway station} (left), \textit{Village} (middle), and \textit{Volleyball court}}. The explanations consist of edges with the top 5\% of attention scores for each example. Our explanations capture local patterns and global interactions closely related to the semantics of the target class. For instance, they identify the connection from the rails and train to the platform to determine a railway station, or from the houses to the mountain to predict a village. The \gls{gsat} method fails to capture these relations, while GNNExplainer also includes many spurious connections.
    \label{fig:qualitative_xai_sun397}}
\end{figure*}

\begin{figure*}[t!]
    \centering
   \makebox[\textwidth]{\textbf{\methodname}}
    \vspace{2mm} 
    \subfloat[Class "Bridge"]{\vspace{0cm}{\includegraphics[width=0.33\textwidth]{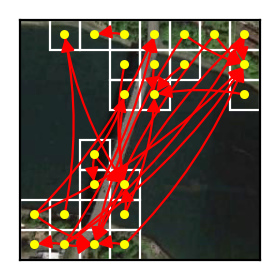}}\label{fig:qualitative_resisc45_bridge}}
    \subfloat[Class "Intersection"]{{\includegraphics[width=0.33\textwidth]{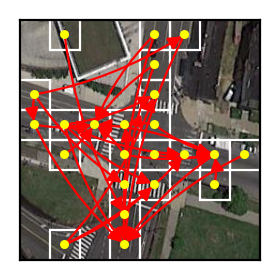}}\label{fig:qualitative_resisc45_intersection}}
    \subfloat[Class "Parking lot"]{
    {\includegraphics[width=0.33\textwidth]{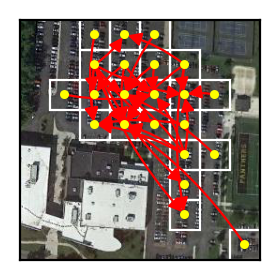}}\label{fig:qualitative_resisc45_parking_lot}}

     \makebox[\textwidth]{\textbf{GNNExplainer}}
    \vspace{-3mm} 

    \subfloat[Class "Bridge"]{\vspace{0cm}{\includegraphics[width=0.33\textwidth]{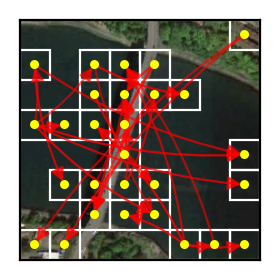}}\label{fig:qualitative_resisc45_gnnexplainer_bridge}}
    \subfloat[Class "Intersection"]{{\includegraphics[width=0.33\textwidth]{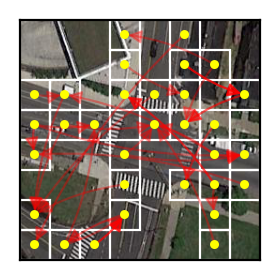}}\label{fig:qualitative_resisc45_gnn_explainer_intersection}}
    \subfloat[Class "Parking lot"]{
        {\includegraphics[width=0.33\textwidth]{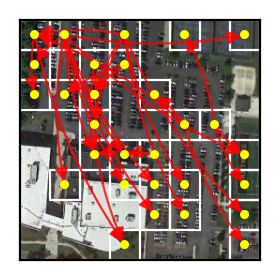} }\label{fig:qualitative_resisc45_gnnexplainer_parking_lot}}

    \makebox[\textwidth]{\textbf{GSAT}}
    \vspace{-3mm} 

    \subfloat[Class "Bridge"]{\vspace{0cm}{\includegraphics[width=0.33\textwidth]{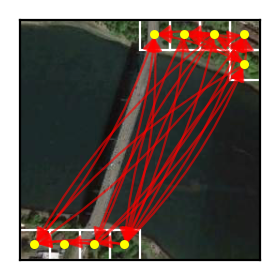}}\label{fig:qualitative_resisc45_gsat_bridge}}
    \subfloat[Class "Intersection"]{{\includegraphics[width=0.33\textwidth]{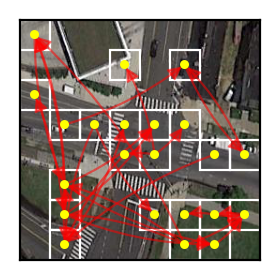}}\label{fig:qualitative_resisc45_gsat_intersection}}
    \subfloat[Class "Parking lot"]{
        {\includegraphics[width=0.33\textwidth]{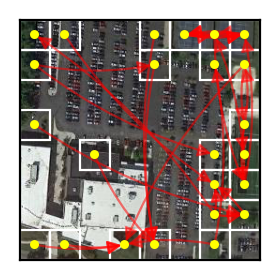} }\label{fig:qualitative_resisc45_gsat_parking_lot}}

    \caption{\textbf{Qualitative comparison of the explanations of our \methodname\ model (top row) against those from \gls{gsat} (bottom row) on the RESISC45 dataset for  classes \textit{Bridge} (left) \textit{Harbor} (middle), \textit{Parking lot} (right).} The explanations consist of edges with the top 5\% of attention scores for each example. The GNNExplainer and \gls{gsat} explanations identify spurious correlations, such as vegetation and buildings for predicting intersections, or sports fields for predicting parking lots. In contrast, our approach on the top row emphasizes the importance of the relations crosswalks and road for intersections, as well as among cars and impervious surfaces for parking lots.}
    \label{fig:qualitative_xai_resisc45}
\end{figure*}

\begin{figure*}[t!]
    \centering
   \makebox[\textwidth]{\textbf{\methodname}}
    \vspace{2mm} 
    \subfloat[Low liveability]{{\includegraphics[width=0.33\textwidth]{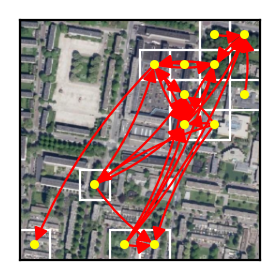}}\label{fig:liveability_low}}
    \subfloat[Medium liveability]{
        {\includegraphics[width=0.33\textwidth]{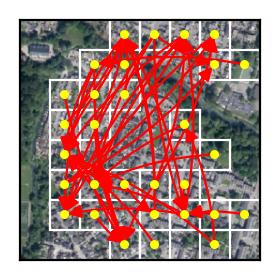} }\label{fig:liveability_medium}}
    \subfloat[High liveability]{\vspace{0cm}{\includegraphics[width=0.33\textwidth]{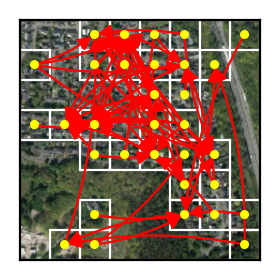}}\label{fig:liveability_high}}

     \makebox[\textwidth]{\textbf{GNNExplainer}}
    \vspace{-3mm} 
    
    \subfloat[Low liveability]{{\includegraphics[width=0.33\textwidth]{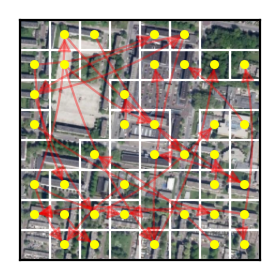}}\label{fig:gnnexplainer_liveability_low}}
    \subfloat[Medium liveability]{
        {\includegraphics[width=0.33\textwidth]{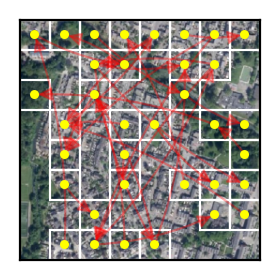} }\label{fig:gnn_explainer_liveability_medium}}
    \subfloat[High liveability]{\vspace{0cm}{\includegraphics[width=0.33\textwidth]{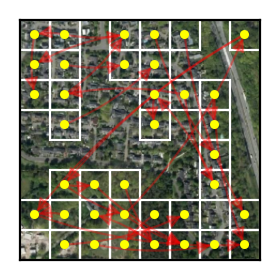}}\label{fig:gnnexplainer_liveability_high}}

    \makebox[\textwidth]{\textbf{GSAT}}
    \vspace{-3mm} 
    
    \subfloat[Low liveability]{{\includegraphics[width=0.33\textwidth]{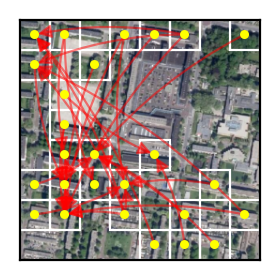}}\label{fig:gsat_liveability_low}}
    \subfloat[Medium liveability]{
        {\includegraphics[width=0.33\textwidth]{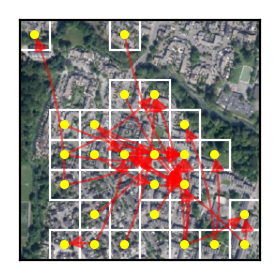} }\label{fig:gsat_liveability_medium}}
    \subfloat[High liveability]{\vspace{0cm}{\includegraphics[width=0.33\textwidth]{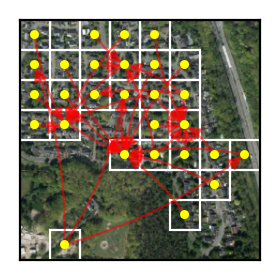}}\label{fig:gsat_liveability_high}}
    
    \caption{\textbf{Qualitative comparison of the explanations of our \methodname\ model (top row) against those from GNNExplainer (middle row) and \gls{gsat} (bottom row) on the Liveability dataset for areas of low liveability (left), medium liveability (middle), and high liveability (right).} The explanations consist of edges with the top 5\% of attention scores for each example.}
    \label{fig:qualitative_xai_liveability}
\end{figure*}

We qualitatively compare the explanations produced by \methodname\ with those derived from GNNExplainer and \gls{gsat} in Figures \ref{fig:qualitative_xai_sun397}, \ref{fig:qualitative_xai_resisc45}, and \ref{fig:qualitative_xai_liveability}.

In detail, Figure \ref{fig:qualitative_xai_sun397} shows explanations for examples from three classes from the SUN397 dataset. First, the explanation subgraph for the \textit{railway station} in the left Figure \ref{fig:qualitative_sun397_railway} reveals that our model identifies local patterns on the train, rails, platforms, windows, and the roof of the station. Equally important, our explanation also includes global edges connecting these components. In contrast, Figures \ref{fig:qualitative_sun397_gnn_explainer_railway} and \ref{fig:qualitative_sun397_gsat_railway} reveal that the most important edges estimated with GNNExplainer mostly focus on interactions between image parts on the roof and the sides, while those from \gls{gsat} capture less semantics with the train station, as they do not identify the rails and train. Further, for the example of the \textit{Village} class in the middle, our model in Figure \ref{fig:qualitative_sun397_village} emphasizes the long-range relations from the houses and the area on the left hill to a few mountain patches on the right. This results in a more compact explanation than GNNExplainer (Figure \ref{fig:qualitative_sun397_gnnexplainer_village}), which emphasizes many interactions between the mountain patches. Further, the \gls{gsat} explanation (Figure \ref{fig:qualitative_sun397_gsat_village}) fails to unveil the long-range relations as it focuses exclusively on the local patterns around the houses on the left hill and on the mountain to the right. Similarly, the explanatory subgraph of our model for the \textit{Volleyball court} class (Figure \ref{fig:qualitative_sun_volleybal_court}) identifies interactions from the players on both sides of the net, as well as from the net to a local connected component to the court ceiling. In contrast, the GNNexplainer subgraph in the middle Figure \ref{fig:qualitative_sun_gnnexplainer_volleybal_court} contains a larger portion of the image with a high focus on interactions on the court floor, while the \gls{gsat} explanation in the bottom Figure \ref{fig:qualitative_sun_gsat_volleybal_court} identifies two disconnected subgraphs: one among the players on the field, and another one to the court ceiling. 

Similarly, the visualized examples from the RESISC45 dataset in Figure \ref{fig:qualitative_xai_resisc45} demonstrate that the relevant edges identified with our sparse attention objective offer enhanced semantics compared to those from GNNExplainer and \gls{gsat}. For instance, Figure \ref{fig:qualitative_resisc45_gsat_bridge} shows an example from the class \textit{Bridge} that reveals that our interpretable \gls{gnn} bottleneck describes a bridge with interactions between the deck, land, and concrete on both sides of the bridge. In contrast, GNNExplainer frequently emphasizes interactions among distant water patches, whereas the \gls{gsat} explanation reveals only interactions between the land. Further, the remaining examples for \textit{Intersection} and \textit{Parking lot} reveal that GNNExplainer and \gls{gsat} explanations can emphasize spurious correlations, such as the relevance of the vegetation and buildings for identifying \textit{intersection} (Figures \ref{fig:qualitative_resisc45_gnn_explainer_intersection} and \ref{fig:qualitative_resisc45_gsat_intersection}) and the vegetation and sport fields around the \textit{parking lot} (Figures \ref{fig:qualitative_resisc45_gnnexplainer_parking_lot} and \ref{fig:qualitative_resisc45_gsat_parking_lot}). On the other hand, the explanatory subgraphs of our method focus on interactions between objects of more related semantics to the target class, highlighting the connections among crosswalks and the road (Figure \ref{fig:qualitative_resisc45_intersection}) for predicting intersection, and among cars and impervious surface for a parking lot (Figure \ref{fig:qualitative_resisc45_parking_lot}).

Finally, the visualized examples from the Liveability dataset in Figure \ref{fig:qualitative_xai_liveability} show that our method produces sparser explanations than GNNExplainer and places greater emphasis on long-range interactions than \gls{gsat}. 

\subsection{Comparison with the \gls{wignet} encoders}
\label{sec:appendix_comparison_encoders}
\begin{figure*}[t!]
    \centering    
    \subfloat[Class "Vilage" (SUN397)]{{\includegraphics[width=0.33\textwidth]{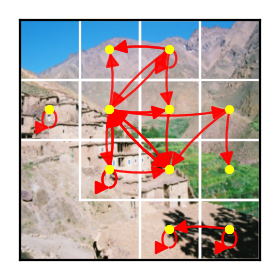}}\label{fig:app_wignn_custom_encoder_sun397}}
    \subfloat[Class "Harbor" (RESISC45)]{
        {\includegraphics[width=0.33\textwidth]{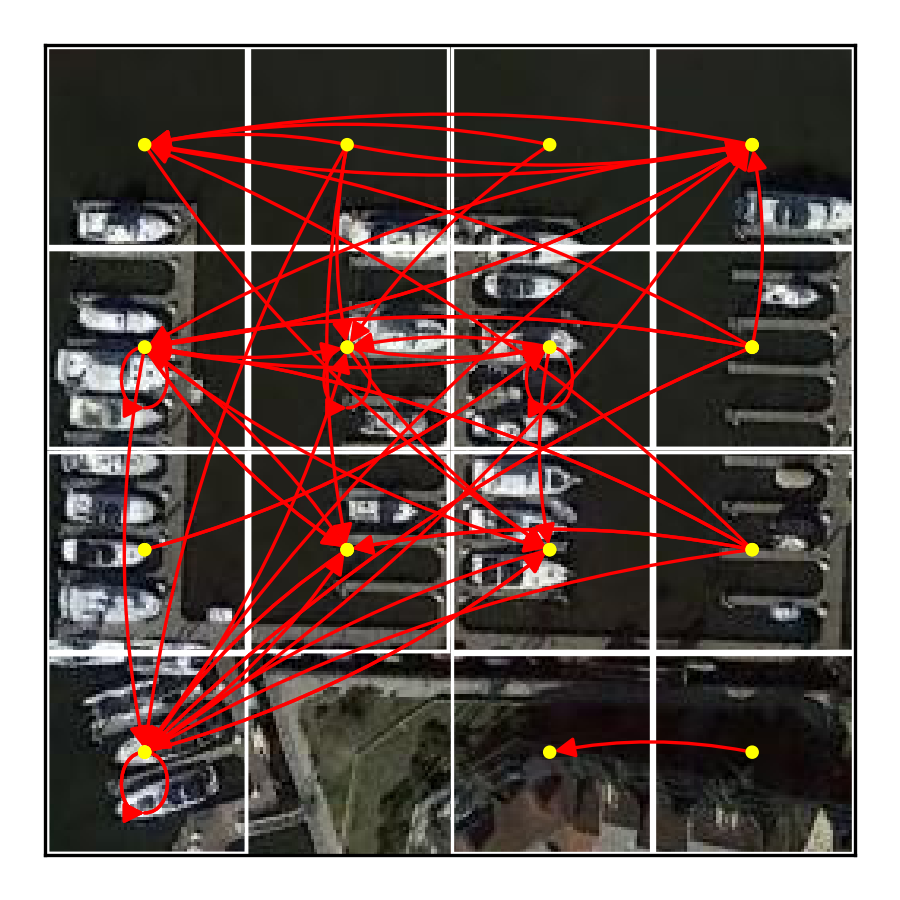} }\label{fig:app_wignn_custom_resisc45}}
    \subfloat["Medium Liveability"]{\vspace{0cm}{\includegraphics[width=0.33\textwidth]{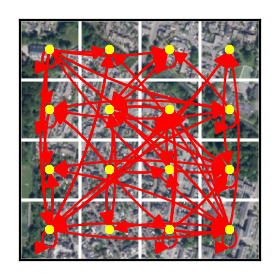}}\label{fig:app_wignn_custom_encoder_liveability}}
    \caption{Visualizing the receptive field of the custom \gls{wignet}  (Conv $\times 2$) encoder for the edges within the top 5\% of attention scores for examples from the class \textit{Village} from the SUN397 dataset  (left), class \textit{Harbor} from the RESISC45 dataset (middle), and medium liveability from the Liveability dataset (right).}
    \label{fig:wignn_conv2_windows}
\end{figure*}

\begin{figure*}[t!]
    \centering    
    \subfloat[Class "Vilage" (SUN397)]{{\includegraphics[width=0.33\textwidth]{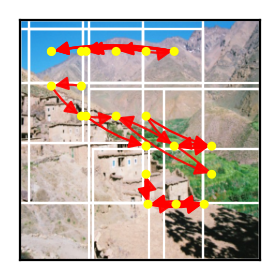}}\label{fig:app_qualitative_sun397}}
    \subfloat[Class "Harbor" (RESISC45)]{
        {\includegraphics[width=0.33\textwidth]{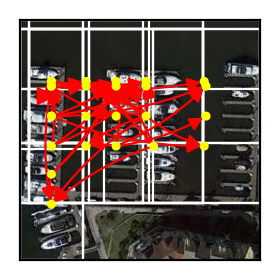} }\label{fig:app_qualitative_resisc45}}
    \subfloat["Medium Liveability"]{\vspace{0cm}{\includegraphics[width=0.33\textwidth]{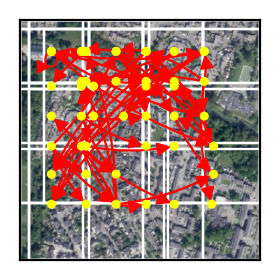}}\label{fig:app_qualitative_liveability}}
    \caption{Visualizing the receptive field of the default \gls{wignet} encoder for the edges within the top 5\% of attention scores for examples from the class \textit{Village} from the SUN397 dataset  (left), class \textit{Harbor} from the RESISC45 dataset (middle), and medium liveability from the Liveability dataset (right).}
    \label{fig:wignn_windows}
\end{figure*}

In Figures \ref{fig:wignn_conv2_windows} and \ref{fig:wignn_windows}, we visualize explanations using the \gls{wignet} encoder with reduced overlap by using Conv $\times 2$ for downsampling as well as the default \gls{wignet} encoder that produces nodes with largely overlapping receptive fields. As Figure \ref{fig:wignn_conv2_windows} demonstrates, the explanations with the  \gls{wignet} (Conv $ \times 2$) encoder contain many self-loops that are an artifact of the encoder that produces redundant nodes that cover the same window in the input image. Further, the nodes in the highlighted subgraphs from the default \gls{wignet} encoder in Figure \ref{fig:wignn_windows} typically cover a large portion of the image, and consequently do not reveal the relevant fine-grained relationships between objects that inform the model's decisions. This is also evident in Table \ref{tab:local_windows_impact} of the main manuscript, which shows lower faithfulness scores for these explanations than for those from our model, which uses a more efficient encoder that creates nodes with smaller, disjoint receptive fields.

\end{document}